%% file: embeddings2.tex
\documentclass{article} % For LaTeX2e
\usepackage{nips14submit_e,times}
\usepackage{amsmath}
\usepackage{amsthm}
\usepackage{amssymb}
\usepackage{graphicx}
\usepackage{xspace}
\usepackage{tabularx}
\usepackage{hyperref}
\usepackage{url}
\usepackage[numbers]{natbib}
\usepackage{wrapfig}
\usepackage[utf8]{inputenc}

\usepackage[normalem]{ulem}

\newcommand{\bmx}[0]{\begin{bmatrix}}
\newcommand{\emx}[0]{\end{bmatrix}}

\graphicspath{ {./figures/} }

\title{Not All Neural Embeddings are Born Equal}

\author{
Felix Hill \\
University of Cambridge
\And
KyungHyun Cho \\
Universit\'{e} de Montr\'{e}al
\And
Sébastien Jean \\
Universit\'{e} de Montr\'{e}al
\And
Coline Devin \\
Harvey Mudd College
\And
Yoshua Bengio \\
Universit\'{e} de Montr\'{e}al, CIFAR Senior Fellow
}

\nipsfinalcopy % Uncomment for camera-ready version

\begin{document}

\maketitle

\input{main.tex}

%\newpage
\bibliography{myref}
\bibliographystyle{plainnat}

% TODO: uncomment if there'll be any appendix
%\newpage
%\appendix
%\input{supp.tex}

\end{document}

%% file: main.tex
\begin{abstract}
Neural language models learn word representations that capture rich linguistic and conceptual information. Here we investigate the embeddings learned by neural machine translation models. We show that translation-based embeddings outperform those learned by cutting-edge monolingual models at single-language tasks requiring knowledge of conceptual similarity and/or syntactic role. The findings suggest that, while monolingual models learn information about how concepts are related, neural-translation models better capture their true ontological status.
\end{abstract}

It is well known that word representations can be learned from the distributional patterns in corpora. Originally, such representations were constructed by counting word co-occurrences, so that the features in one word's representation corresponded to other words~\citep{landauer1997solution,turney2010frequency}. Neural language models, an alternative means to learn word representations, use language data to optimise (latent) features with respect to a language modelling objective. The objective can be to predict either the next word given the initial words of a sentence~\citep{Bengio2003lm,mnih2009scalable,collobert2008unified}, or simply a nearby word given a single cue word~\citep{mikolov2013distributed,Pennington2014}. The representations learned by neural models (sometimes called \emph{embeddings}) generally outperform those acquired by co-occurrence counting models when applied to NLP tasks~\citep{baroni2014don}. 

Despite these clear results, it is not well understood how the architecture of neural models affects the information encoded in their embeddings. Here, we explore this question by considering the embeddings learned by architectures with a very different objective function to monolingual language models: \emph{neural machine translation models}. We show that translation-based embeddings outperform monolingual embeddings on two types of task: those that require knowledge of conceptual similarity (rather than simply association or relatedness), and those that require knowledge of syntactic role. We discuss what the findings indicate about the information content of different embeddings, and suggest how this content might emerge as a consequence of the translation objective. 

% We show that the precise information encoFinally, we consider ways to combine the distinct information encoded in language-model embeddings and those from translation embeddings. And we consider ways to overcome two important limitations of neural translation models as embedding-learning architectures: the computational difficulty in learning large vocabularies of embeddings, and the requirement for sentence-aligned bilingual training corpora.   

\section{Learning embeddings from language data}

Both neural language models and translation models learn real-valued embeddings (of specified dimension) for words in some pre-specified vocabulary, \(V\), covering many or all words in their training corpus. At each training step, a `score' for the current training example (or batch) is computed based on the embeddings in their current state. This score is compared to the model's objective function, and the error is backpropagated to update both the model weights (affecting how the score is computed from the embeddings) and the embedding features. At the end of this process, the embeddings should encode information that enables the model to optimally satisfy its objective.     
\subsection{Monolingual models}

In the original neural language model~\cite{Bengio2003lm} and subsequent
variants \cite{collobert2008unified}, each training example consists of $n$ subsequent words, of
which the model is trained to predict the $n$-th word given the first $n-1$
words. The model first represents the input as an ordered sequence of embeddings,
which it transforms into a single fixed length `hidden' representation
by, e.g., concatenation and non-linear projection.
Based on this representation, a probability distribution is
computed over the vocabulary, from which the model can sample a guess at the
next word. The model weights and embeddings are updated to maximise the
probability of correct guesses for all sentences in the
training corpus. 
 
More recent work has shown that high quality word embeddings can be learned via models
with no nonlinear hidden layer ~\citep{mikolov2013distributed,Pennington2014}. Given a single word in
the corpus, these models simply predict which other words will occur nearby. For each
word \( w\) in \(V\), a list of training cases \({(w,c) : c \in V }\) is
extracted from the training corpus.
%according to some algorithm. 
For instance, in the \emph{skipgram}
approach ~\citep{mikolov2013distributed}, for each `cue word' \(w\) the `context words' \(c\) are sampled
from windows either side of tokens of \(w\) in the corpus (with \(c\) more
likely to be sampled if it occurs closer to \(w\)).\footnote{
    Subsequent variants use different algorithms for selecting the
    \((w,c)\) from the training corpus \citep{Hill2014EMNLP,levy2014dependency}
} For each \(w\) in \( V\), the model initialises both a
cue-embedding, representing the \(w\) when it occurs as a cue-word, and a
context-embedding, used when \(w\) occurs as a context-word. For a cue word
\(w\), the model can use the corresponding cue-embedding and all
context-embeddings to compute a probability distribution over \(V\) that
reflects the probability of a word occurring in the context of \(w\). When a
training example \((w,c)\) is observed, the model updates both the cue-word
embedding of \(w\) and the context-word embeddings in order to increase the
conditional probability of \(c\). 

\subsection{Translation-based embeddings}

Neural translation models generate an appropriate sentence in their target
language \(S_t\)  given a sentence \(S_s\) in their source language~\citep[see,
e.g.,][]{Sutskever2014sequence,Cho2014a}. In doing so, they learn distinct sets of
embeddings for the vocabularies \(V_ s\) and \(V_t\) in the source and target
languages respectively.

Observing a training case \((S_s, S_t)\), such a model represents \(S_s\) as an ordered sequence of embeddings of words from \(V_s\). The sequence for \(S_s\) is then encoded into a single representation \(R_S\).\footnote{Alternatively, subsequences (phrases) of \(S_s\) may be encoded at this stage in place of the whole sentence~\citep{Bahdanau2014}.} Finally, by referencing the embeddings in \(V_t\), \(R_S\) and a representation of what has been generated thus far, the model decodes a sentence in the target language word by word. If at any stage the decoded word does not match the corresponding word in the training target \(S_t\), the error is recorded. The weights and embeddings in the model, which together parameterise the encoding and decoding process, are updated based on the accumulated error once the sentence decoding is complete. 

Although neural translation models can differ in low-level architecture~\citep{Cho2014,Bahdanau2014}, the translation objective exerts similar pressure on the embeddings in all cases. The source language embeddings must be such that the model can combine them to form single representations for ordered sequences of multiple words (which in turn must enable the decoding process). The target language embeddings must facilitate the process of decoding these representations into correct target-language sentences.

\section{Comparing Mono-lingual and Translation-based Embeddings}

To learn translation-based embeddings, we trained both the RNN encoder-decoder ~\citep[\emph{RNNenc},][]{Cho2014} and the \emph{RNN Search} architectures~\citep{Bahdanau2014} on a 300m word corpus of English-French sentence pairs. We conducted all experiments with the resulting (English) source embeddings from these models. For comparison, we trained a monolingual skipgram model~\citep{mikolov2013distributed} and its \emph{Glove} variant~\citep{Pennington2014} for the same number of epochs on the English half of the bilingual corpus. We also extracted embeddings from a full-sentence language model~\citep[\emph{CW},][]{collobert2008unified} trained for several months on a larger 1bn word corpus. 
  
As in previous studies~\citep{Agirre2009,Bruni2014,baroni2014don}, we evaluate embeddings by calculating pairwise (cosine) distances and correlating these distances with (gold-standard) human judgements.    
Table~\ref{table:perf} shows the correlations of different model embeddings with three such gold-standard resources, WordSim-353~\citep{Agirre2009}, MEN~\citep{Bruni2014} and SimLex-999~\citep{hill2014simlex}. Interestingly,
translation embeddings perform best on SimLex-999, while the two sets of monolingual embeddings perform better on modelling the MEN and WordSim-353. To interpret these results, it should be noted that SimLex-999 evaluation quantifies conceptual \emph{similarity} (\emph{dog} - \emph{wolf}), whereas MEN and WordSim-353 (despite its name) quantify more general \emph{relatedness} (\emph{dog} - \emph{collar})~\citep{hill2014simlex}. The results seem to indicate that translation-based embeddings better capture similarity, while monolingual embeddings better capture relatedness. 

\begin{table}[t]
\begin{center}
\begin{tabular}{r c | r  r  r  r r}
    \multicolumn{2}{c|}{~} &\bf Skipgram &\bf Glove &\bf CW &\bf RNNenc &\bf Search \\ 
\hline
WordSim-353   & \(\rho\) & 0.52 & 0.55 & 0.51 & 0.57 & { \bf 0.58} \\
MEN & \(\rho\) & 0.44 & {\bf 0.71} & 0.60 & 0.63 &  0.62  \\
SimLex-999 & \(\rho\) & 0.29 & 0.32 & 0.28 & {\bf 0.52} &  0.49 \\
\hline

TOEFL & \(\%\) & 0.75 & 0.78 & 0.64 & {\bf 0.93} &  {\bf 0.93} \\
Syn/antonym & \(\%\) & 0.69  & 0.72  &  0.75 & {\bf 0.79} & 0.74 \\
\hline
\emph{teacher} & nn & {\small \emph{vocational}} &  {\small \emph{student}} 
& {\small \emph{student}} & {\small \emph{professor}}  & {\small \emph{instructor}} \\ 
\emph{white} & nn & {\small \emph{red}} &  {\small \emph{red}} 
&  {\small \emph{black}} & {\small \emph{blank}} & {\small \emph{black}} \\ 
\emph{heat} & nn & {\small \emph{thermal}} &  {\small \emph{thermal}} 
& {\small \emph{wind}} & {\small  \emph{warmth}} & {\small \emph{warmth}} \\ 
\end{tabular}
\caption{Translation-based embeddings outperform alternatives on similarity-focused evaluations.}
\label{table:perf}
\end{center}
\vspace{-5mm}
\end{table}

To test this hypothesis further, we ran two more evaluations focused specifically on similarity. The TOEFL synonym test contains 80 cue words, each with four possible answers, of which one is a correct synonym~\citep{landauer1997solution}. We computed the proportion of questions answered correctly by each model, where a model's answer was the nearest (cosine) neighbour to the cue word in its vocabulary.\footnote{To control for different vocabularies, we restricted the effective vocabulary of each model to the intersection of all model vocabularies, and excluded all questions that contained an answer outiside of this intersection.} In addition, we tested how well different embeddings enabled a supervised classifier to distinguish between synonyms and antonyms. For 500 hand-labelled pairs we presented a Gaussian SVM with the concatenation of the two word embeddings. We evaluated accuracy using 8-fold cross-validation. 

As shown in Table~\ref{table:perf}, translation-based embeddings outperform
all monolingual embeddings on these two additional similarity-focused tasks. Qualitative analysis of nearest neighbours (bottom rows) also supports the conclusion that proximity in the
translation embedding space corresponds to similarity while proximity in the monolingual embedding space reflects relatedness.

%\footnote{Readers can inspect nearest neighbours in each embedding space using our web demo.} 
 
\subsection{Quantity of training data}

\begin{figure*}[h]
\includegraphics[width = \textwidth,clip=True,trim=0 10 0 10]{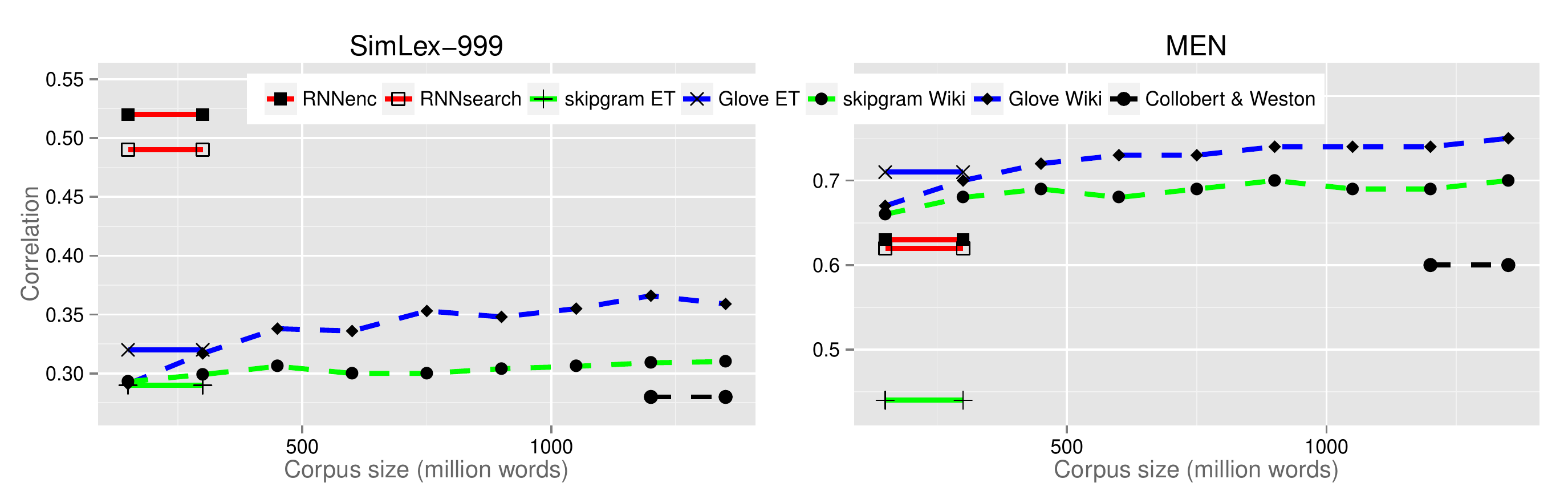}
\vspace{-4mm}
\caption{Effect of training corpus size on performance. WordSim-353 results were similar to MEN.}
\label{fig:size}
\end{figure*}

In previous work, monolingual models were trained on corpora many
times larger than the English half of our parallel translation corpus. To check if these models simply need more training data to
capture similarity as effectively as translation models, we trained them on
increasingly large subsets of Wikipedia.\footnote{
    We could not do the same for the translation models because of the scarcity of
    bilingual corpora.
} The results refute this possibility: the performance of monolingual embeddings on similarity tasks converges
well below the level of the translation-based embeddings (Fig.~\ref{fig:size}).

\subsection{Analogy questions}

Lexical analogy questions are an alternative way of evaluating word representations~\citep{mikolov2013distributed,Pennington2014}. In this task, models must identify the correct answer (\emph{girl}) when
presented with questions such as `\emph{man} is to \emph{boy} as \emph{woman} is to ...'. For skipgram-style embeddings, it has been shown that if \( \bf m, b \) and \( \bf w\) are the embeddings for \emph{man}, \emph{boy} and \emph{woman} respectively, the correct answer is often the nearest neighbour in the vocabulary (by cosine distance) to the vector \( \bf v = w + b - m \)~\citep{mikolov2013distributed}. 

\begin{figure*}[t]

\includegraphics[width = \textwidth,clip=True,trim=0 10 0 10]{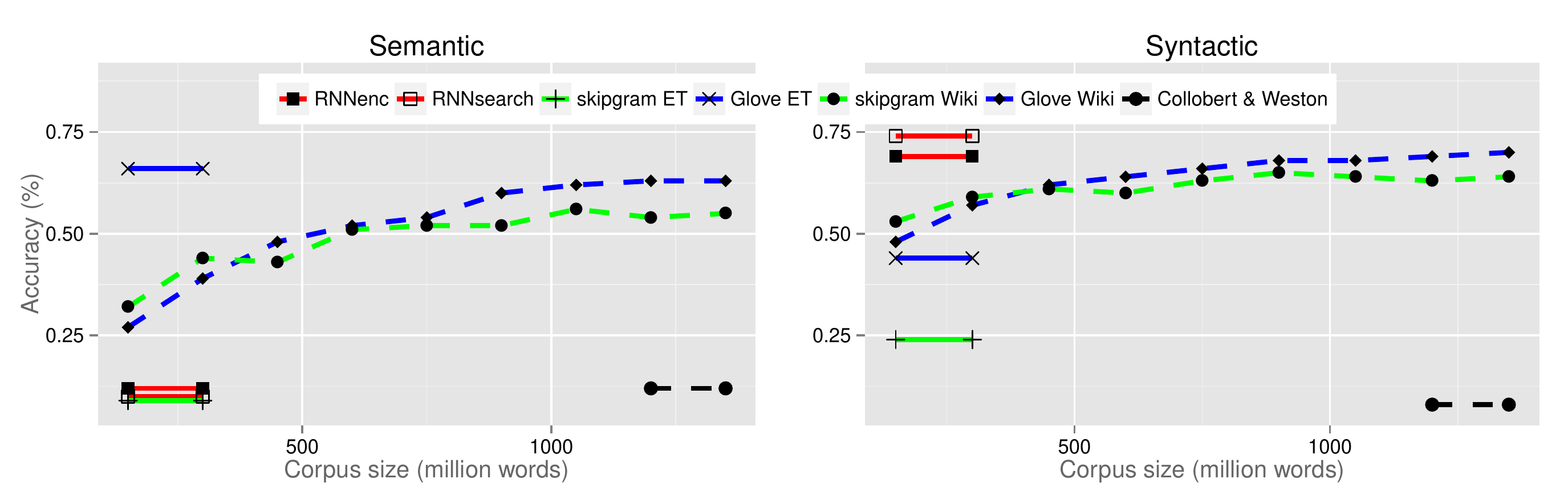}

\vspace{-4mm}
\caption{Translation-based embeddings perform best on syntactic analogies (\emph{run,ran: hide, hid}). Monolingual skipgram/Glove models are better at semantic analogies (\emph{father, man; mother, woman})}
\label{fig:analogy}

\end{figure*}

We evaluated the embeddings on this task using the same vector-algebra method as~\citep{mikolov2013distributed}. As before we excluded questions containing a word outside the intersection of all model vocabularies, and restricted all answer searches to this reduced
vocabulary, leaving 11,166 analogies. Of these, 7219 are classed as `syntactic', in that they exemplify mappings between parts-of-speech or syntactic roles (\emph{fast, fastest; heavy | heaviest}), and 3947 are classed as `semantic` (\emph{Ottawa, Canada; Paris | France}), deriving from wider world knowledge. As shown in Fig.~\ref{fig:analogy}, the translation-based embeddings seem to yield poor answers to semantic analogy questions, but are very effective for syntactic analogies, outperforming the monolingual embeddings, even those trained on much more data.

\section{Conclusions}

Neural machine translation models are more effective than monolingual models at learning embeddings that encode information about concept similarity and syntactic role. In contrast, monolingual models encode 
general inter-concept relatedness (as applicable to semantic analogy questions), but struggle to capture similarity, even when training on larger corpora. For skipgram-style models, whose objective is to predict linguistically collocated pairs, this limitation is perhaps unsurprising, since co-occurring words are, in general, neither semantically nor syntactically similar. However, the fact that it also applies to the full-sentence model \emph{CW} suggests that inferring similarity is problematic for monolingual models even with knowledge of the precise (ordered) contexts of words. This may be because very dissimilar words (such as antonyms) actually often occur in identical linguistic contexts. 

When considering the strengths of translation embeddings - similarity and syntactic role - it is notable that each item in the three similarity-focused evaluations consists of word groups or pairs of identical syntactic role. Thus, the strong performance of translation embeddings on similarity tasks cannot be simply a result of their encoding of richer syntactic information. To perform well on SimLex-999, embeddings must encode information approximating what concepts \emph{are} (their function or ontology), even when this contradicts the signal conferred by co-occurrence (as can be the case for related-but-dissimilar concept pairs)~\citep{hill2014simlex}. The translation objective seems particularly effective at inducing models to encode such ontological or functional information in word embeddings. 

% Indeed, while the selection of training examples based on dependency parses
% rather than standard bag-of-words input can improve the ability of
% mono-lingual models to capture syntactic roles [REF], the improvement in
% similarity modelling generated by this modification is far smaller than that
% produced by translation embeddings [].  

While much remains unknown about this process, one cause might be the different ways in which words partition the meaning space of a language. In cases where a French word has two possible English translations (e.g. \emph{gagner \(\to\) win} / \emph{earn}), we note that the (source) embeddings of the two English words are very close. It appears that, since the translation model, which has limited encoding capacity, is trained to map tokens of \emph{win} and \emph{earn} to the same place in the target embedding space, it is efficient to move these concepts closer in the source space. While clear-cut differences in how languages partition meaning space, such as (\emph{gagner = win, earn}), may in fact be detrimental to similarity modelling (\emph{win} and \emph{earn} are not synonymous to English speakers), in general, languages partition meaning space in less drastically different ways. We hypothesize that these small differences are the key to how neural translation models approximate ontological similarity so effectively. At the same time, since two dissimilar or even antonymous words in the source language should never correspond to a single word in the target language, these pairs diverge in the embedding space, rendering two antonymous embeddings easily distinguishable from those of two synonyms.   

%\section*{Acknowledgments}
%
%The authors would like to thank the developers of
%Theano~\cite{bergstra+al:2010-scipy,Bastien-Theano-2012}.  We acknowledge the
%support of the following agencies for research funding and computing support:
%NSERC, Calcul Qu\'{e}bec, Compute Canada, the Canada Research Chairs and CIFAR.